# Causal machine learning for single-cell genomics

Alejandro Tejada-Lapuerta[1,5]*, Paul Bertin[2,3]*, Stefan Bauer[4,5], Hananeh Aliee[6 †], Yoshua Bengio[2,3 †], Fabian J. Theis[1,5 †]

[1] Institute of Computational Biology, Helmholtz Munich, Munich, Germany
[2] Mila, the Quebec AI Institute, Montreal, Canada
[3] Université de Montréal, Montréal, Canada
[4] Helmholtz AI, Munich, Germany
[5] Technical University of Munich, Munich, Germany
[6] Sanger Institute, Cambridge, UK
* Equal contribution
† Correspondence to < ha10@sanger.ac.uk >, < yoshua.bengio@mila.quebec > and < fabian.theis@helmholtz-munich.de >

**Abstract**

Advances in single-cell omics allow for unprecedented insights into the transcription profiles of individual cells. When combined with large-scale perturbation screens, through which specific biological mechanisms can be targeted, these technologies allow for measuring the effect of targeted perturbations on the whole transcriptome. These advances provide an opportunity to better understand the causative role of genes in complex biological processes such as gene regulation, disease progression or cellular development. However, the high-dimensional nature of the data, coupled with the intricate complexity of biological systems renders this task nontrivial. Within the machine learning community, there has been a recent increase of interest in causality, with a focus on adapting established causal techniques and algorithms to handle high-dimensional data. In this perspective, we delineate the application of these methodologies within the realm of single-cell genomics and their challenges. We first present the model that underlies most of current causal approaches to single-cell biology and discuss and challenge the assumptions it entails from the biological point of view. We then identify open problems in the application of causal approaches to single-cell data: generalising to unseen environments, learning interpretable models, and learning causal models of dynamics. For each problem, we discuss how various research directions – including the development of computational approaches and the adaptation of experimental protocols – may offer ways forward, or on the contrary pose some difficulties. With the advent of single cell atlases and increasing perturbation data, we expect causal models to become a crucial tool for informed experimental design.

# Introduction and motivation

Cells are the basic unit of life, and the biological functions they perform, as well as the identity they acquire are the result of physical and biochemical processes that happen in nature. In particular, these processes influence how cells respond to treatments and other perturbations. Technological advances in molecular profiling at single-cell resolution have provided an unprecedented view on cellular processes. It is now possible to jointly measure multiple modalities, such as chromatin accessibility, RNA expression and protein abundance, in a single cell[1–4]. Furthermore, such rich molecular measurements can be accompanied by additional information on the temporal or spatial context of the cell[5–9]. The above measurements can be performed at high-throughput, allowing us to sample several thousands of single cells in a single experiment – and millions of cells by a single lab within weeks –, enabling the assembly of detailed molecular maps of cellular variation and tissue organisation. Classical scientific approaches may fail at modelling and identifying the behaviours and dynamics of complex biological systems due to the huge number of variables of interest – typically in the order of $10^4$ to $10^5$ –[1*], the often multimodal nature of the biological observations and partial observability of the system.

Machine learning has proven useful for solving complex tasks on high dimensional data in fields such as computer vision and natural language processing. Advances in machine learning and artificial intelligence have now permeated into the field of single cell genomics, with crucial contributions such as dimensionality reduction and visualisation[11,12], data integration[13], trajectory inference[14] and transfer across modalities[15]. However, progress towards a better understanding of the underlying biological processes and the quantitative prediction of experimental outcomes have for now been limited. The large majority of the machine learning methods applied to single cell genomics have their foundations in statistical learning, and are designed to find and leverage patterns in a single, fixed, data distribution. Nevertheless, when the experimental conditions change – due to either external conditions, some perturbation, or a small variation in the subject of study – these patterns may not be relevant anymore, and statistical learning may fail to generalise[16]. Statistical learning's limitations in generalisation arise from its dependence on established data patterns, which can hinder its adaptability to accommodate shifts, novel influences, and complex dynamics under new experimental conditions, for example under the putative effect of drugs.

In contrast, causal learning approaches model a family of distributions[17], wherein each distribution is associated with a so-called *environment*, that could typically correspond to a specific experimental condition, genetic perturbation or a different experimental exposure (disease progression, developmental step, pathogen, drug, chemical, radiation…). Causal learning approaches aim at capturing interactions which remain unchanged across environments between the variables that underlie the data generative process. This is in line

---

[1*] In human cells, there are ~20k genes, and ~70k proteins, taking splicing variants into account[10].

with one of the key aspects of scientific endeavours, and biological research in particular, where scientists want to discover relationships between entities, or laws, which are stable across environments and not uniquely limited to the set of data under study. The ability to model unseen environments would open new possibilities for guiding scientific discovery using computational approaches[18]. Moreover, causal learning approaches often improve interpretability when compared to standard statistical approaches[19], which is of the utmost importance for explicitly discovering novel biological mechanisms or relationships. In the context of cell biology, interpretability could help provide human-understandable explanations for the predictions of the model and point to the causes that lead to a specific cellular response. This would be substantial for designing subsequent experiments targeted towards the validation of the explanations provided by the model.

In the machine learning community, causality has enjoyed a renewed interest, and many methods, algorithms and novel theoretical paradigms are emerging. For instance the recent resurgence of interest in causal machine learning has proposed adapting frameworks such as Structural Causal Models[17] (SCM) to high dimensional data, thereby opening the way to applications of these methods in transcriptomics. This framework can be used in many contexts, including causal inference (i.e. measuring the strength of an interaction) and causal discovery (i.e. discovering which variables interact). Yet, in order for causal approaches to make an impact on cell biology, it is imperative that they be adapted to the specificities of single cell transcriptomics – such as the dynamic nature of gene interactions, data biases, partial observability and lack of ground truth. The objective of this work is to identify and analyse open problems faced by the field, as well as putting them into perspective with ongoing research directions.

After providing a brief history of and background on causal inference techniques in genetics and transcriptomics, we conceptualise and review the model that underlies most of current causal approaches to single-cell biology – which we call the *default cellSCM*. We analyse this model in depth by discussing the assumptions it entails from the biological point of view and highlight its limitations dealing with important aspects of single-cell transcriptomics. We then discuss three open problems: generalisation to unseen environments, interpretability of models and modelization of the dynamical aspect of biological mechanisms. For each problem, we discuss how various research directions – including the development of computational approaches and the adaptation of experimental protocols – may offer ways forward, or on the contrary pose some difficulties.

# Background

Causality has always been a fundamental question that has motivated biological research at every level. The relationship between a genetic variant and a disease phenotype, the efficacy of a drug in a population, or the effect of a genetic perturbation in a cell are causal questions in the sense that they focus on the effect of a cause (the mutation, drug, or genetic perturbation in the examples above). Causal inference has been largely applied to genome-wide association studies (GWAS) to discover the relationships between different genetic variants and disease phenotypes, employing approaches like Mendelian Randomization[20] or linkage disequilibrium score regression[21].

In genomics, there has been a long interest in discovering interactions between bioentities such as genes that could provide mechanistic explanations of biological processes, like cellular dynamics during development or disease progression. One may seek to discover collaborative assemblies of biological entities. For instance, traditional approaches typically attempt to discover *module networks* by finding small groups of genes (a.k.a. modules) that function together and whose expressions are tightly correlated[22]. However, most causality-related approaches in genomics focus on the task of Gene Regulatory Network (GRN) inference, where directed connections going from a regulator to a regulated gene are learned. Those approaches range from the use of conditional independence testing[23,24] to Granger causality[25], as well as amortised approaches based on supervised learning[26]. An example of a conditional independence based method is the PC algorithm[27]. It determines whether two variables (in this case, two genes) are independent given a set of other variables (other genes) employing statistical tests. If the statistical test shows that two variables are independent given a set of other variables, it means that there is no direct causal connection between the two variables. When analysing time series, Granger causality offers a distinct interpretation of causality. It states that a variable serves as a cause for another variable if its value at a particular time step affects the value of the latter variable at the following time step. Still, a large line of research tackles the task of GRN inference while relying on statistical learning only, and tries to incorporate and leverage the specificities of transcriptomic data to improve the inference[28]. These approaches usually operate with heuristics using multimodal data (mainly RNA-seq and ATAC-seq)[29,30], prior knowledge – such as known transcription factor's (TF) binding sites – and an extensive use of bioinformatic analysis tools.

For all methods mentioned above, the experimental validation of the inferred GRNs has been a huge challenge[28]. Indeed, the true GRN is mostly unknown in the case of human cells. Other organisms, s.a. E. Coli are better understood, and regulation databases exist[100], but they are still noisy and incomplete. However, the abundance – expected to grow further in the upcoming years – of perturbational data may offer a way forward, both for validating GRNs and for the applicability of causal approaches to transcriptomics. Indeed, interventions are a crucial tool for causal inference, and observations of how the system reacts to different perturbations provides valuable information on the mechanisms at play. Usually, identifying the true interactions from observational data alone is not possible (without specific assumptions). Hence, several causal structure learning methods rely on interventional data to learn the causal mechanisms[31–34].

# Causality for single-cell genomics

In this section, we begin by proposing a definition of causal models – slightly different from the usual one but which will be useful later in the discussion – as a pair of a graph-based model and an interventional model. We present how Structural Causal Models (SCMs) – a common example of graph-based models – are typically applied to transcriptomics and call this model the *default cellSCM.* We then describe a simple interventional model that is often used to model CRISPR knockout perturbations, and sometimes drug perturbations, as well as a more advanced interventional model. All models presented in this section obviously correspond to a simplified view of the cell, and will serve as the starting point for the rest of our discussion. For each component of the *default cellSCM*, we examine what its assumptions entail from the biological point of view and whether we can reasonably expect them to hold in the context of transcriptomics.

**A definition of causal models**

---

**Box 1 Causal models**

**Causal model**

A model able to generate an entire family of distributions, each distribution corresponds to a different environment (e.g. experimental condition). It is usually a pair $(g, h)$ composed of:

- A **graph-based model** $g_\theta$ which encodes explicit relationships between causal variables (associated with the nodes of the graph) and is a generative model over the space X (typically gene expression space, $R^{20k}_+$) with parameters $\theta \in \Theta$. Parameters $\theta$ typically include the adjacency matrix of the graph.

- An **interventional model** $h: I \times \Theta \to \Theta$ which modifies the graph-based model such that it can generate samples from interventional distributions. Given an intervention $i \in I$ and the current (unintervened) parameters $\theta \in \Theta$ of the graph-based model, $h$ generates intervened parameters $\tilde{\theta}$ such that $g_{\tilde{\theta}}$ generates samples from the interventional distribution associated with $i$. Typically, $h$ modifies the adjacency matrix of the graph-based model and removes some edges.

**Structural Causal Model (SCM)**[35]

A type of graph based model wherein the value of each variable is generated through a so-called structural assignment which takes the value of its parents as input:

$$X_i := f_i(PA_i, U_i), (i = 1, ..., n)$$

where $f_i$ is a deterministic function, $PA_i \subset \{X_1, ..., X_n\} \setminus X_i$ is the set of causal parents of node $X_i$ in the graph, and $U_i$ is a random variable that represents the variability (within the population of cells) about the generative process of $X_i$ given its parents $PA_i$. The set of noise variables $U_1, ..., U_n$ are jointly independent. Most importantly, the graph is required to be acyclic, in order to be able to sample from the joint distribution via ancestral sampling.

---

> **Dynamic Structural Causal Model[36]**
>
> A type of graph based model wherein structural assignments govern the temporal evolution of causal variables:
>
> $$dX_i := f_i(PA_i, X_i)dt + \sigma(PA_i, X_i)d\epsilon_i, \ (i = 1, ..., n)$$
>
> where $d\epsilon_i$ is a Wiener ("white noise") process. In the case where time evolution is assumed to be deterministic, we get structural assignments of the form:
>
> $$\frac{dX_i}{dt} = f_i(PA_i, X_i), \ (i = 1, ..., n)$$
>
> The functions $\{f_i\}_i$ can be referred to collectively as the evolution function of the system.
>
> **Intervention**
>
> An intervention is any action performed by the interventional model over the set of parameters of the graph based model. An intervention is called perfect if it removes all incoming edges of a specific causal variable $X_i$, which is said to be *intervened on*. This removes the dependency of $X_i$ on its causal parents, and $X_i$ is set to a given value. Likewise, an intervention is called imperfect if the parameters of the structural assignment between the intervened variable and its causal parents are modified but dependency on parents is not removed.

The rules that underlie biological processes can be represented within a so-called causal model, used to generate observations of cells across different environments. Here, the environment can be thought of as all the relevant aspects of the cell population on which the experiment is performed, as well as the experimental protocol (i.e. the list of instructions a lab experimentalist has to follow in order to perform the experiment: grow the cells, potentially apply some perturbation, perform sequencing...). This definition of the *environment* corresponds to the terminology used in the causality community and is broader than the notion of *extracellular environment* used in cell biology. In single-cell genomics, we can distinguish between three types of environmental aspects: biological perturbations (mainly small molecules and CRISPR knockouts), pure biological aspects of the population of cells (related to cellular identity, s.a. cell line or mutation information) and technical aspects (e.g. incubation time, media). In what follows, we will not rigorously distinguish between these different types of distributional shifts and will consider that each environment corresponds to the application of an intervention.

As each single-cell experiment usually involves a large population of cells, the outcome of an experiment is usually represented as a distribution over cellular states. We define a causal model as a model that is able to generate an entire family of distributions, wherein each distribution corresponds to a different environment. Although not strictly necessary, most causal models consist of a pair of a graph-based model, that encodes explicit relationships between causal variables (associated with the nodes of the graph), and an interventional model that, given an intervention, modifies the graph-based model such that it can generate the interventional distributions. In that case, the graph-based model is a generative model capable of sampling from the observational distribution or from any interventional distribution if modified by the interventional model.

**Graph-based model of a cell: the *default cellSCM***

Here, we introduce the model that underlies most of current causal approaches to single-cell biology[24,37–39], which we will call the *default cellSCM*. We model the physical generative process that controls gene expression as a Structural Causal Model (SCM) where the causal variables are the biological variables of interest, in this case assumed to be the expression levels (i.e. total amount of RNA transcripts within the cell) of a set of genes. Causal relationships between variables are encoded through a graph whose edges point from parent variables to child variables. The value of each variable is generated through a so-called structural assignment which takes the value of its parents as input, and some noise that represents variability between cells. The graph is required to be acyclic, in order to be able to sample from the joint distribution via ancestral sampling. The *default cellSCM* model is represented in Figure 1.A.

While time is abstracted away in the *default cellSCM*, it is useful to remind ourselves that its causal variables actually correspond to the expression levels at the time of acquisition $t_{acq}$, which is the only time at which expressions are observed. Therefore, if one were to explicitly refer to time, the *default cellSCM* would assume that the expression levels of genes G1 and G3 at time $t_{acq}$ regulate the expression level of G2 at time $t_{acq}$. This choice is primarily motivated by practical considerations – it is convenient to model relationships between observed quantities.

To summarise, under the *default cellSCM* assumption, observed gene expressions are generated through the application of the structural assignments of the SCM, and the noise variables model the variability between cells. However, this model is a simplified view of the true biological processes at play, and has several limitations that we cover in detail in the remainder of this section. In short, this model does not account for any unobserved variable, does not take time into account, and is unable to capture cyclic interactions between genes. Last but not least, we show that one of the chore assumptions of the model, the independence of mechanisms governing expression levels, does not hold.

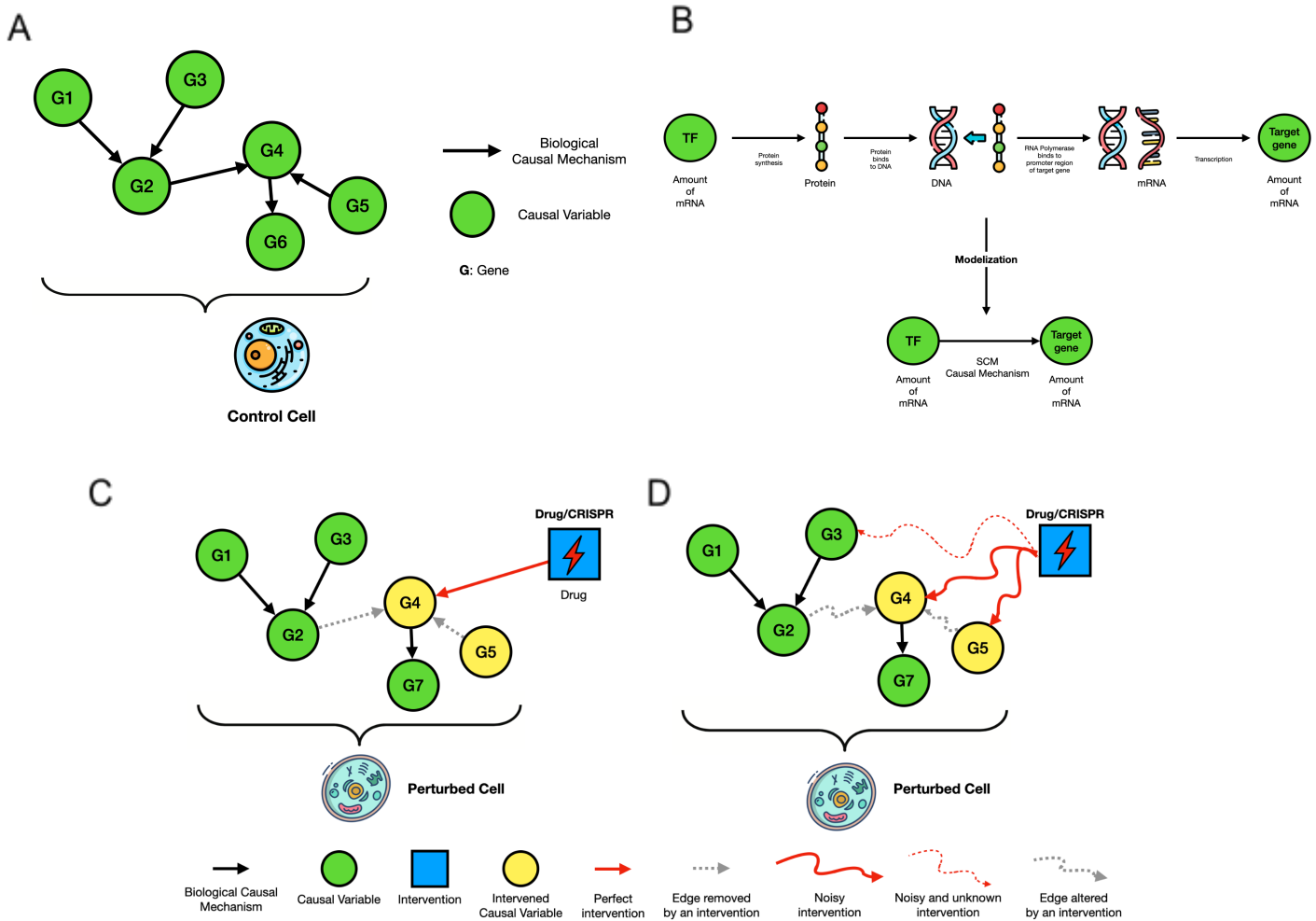

**Figure.1:** Overview of the default cellSCM model **(A)** Modelling of a cell as a SCM through a model we call the *default cellSCM*. Edges represent the causal relationships between the genes. In the absence of any intervention over the cell, the model generates expression profiles associated with control cells. Genes G1 and G3 are causal parents of G2, so there exists a structural assignment through which the value of G2 is computed given the values of G1 and G3, while G1 and G3 would be master regulator genes whose expression is not affected by other genes. Note that the noise random variables are not represented explicitly on this diagram **(B)** The causal mechanisms of an SCM do not capture all the complexity of the true biological processes but instead capture their distributional properties. **(C)** In this simplified model, the perturbations are naively treated as perfect interventions where dependence of targeted variables on their regulators is completely removed. The expression level of gene G4 is directly set by the intervention, it does not depend on the expressions of G2 and G5 anymore. **(D)** A more accurate way to model biological perturbations is to allow for several targets, to account for uncertainties in the interventional targets, and to not fully remove dependency on regulators.

**Abstracting away some of the complexity of biological mechanisms**

It is important to note that capturing the whole complexity of the true physical process is unfeasible. Instead, the *default cellSCM* presented above captures the real biological mechanisms in a simplified way. For instance, as depicted in Figure 1.B, the regulation of the expression of a target gene by a TF-coding gene involves a complex chemical process in which the TF protein synthesised by the regulating gene binds a specific region of the DNA, easing or hindering the binding of the RNA polymerase – the protein responsible for transcription – to the promoter region of the regulated gene. In single-cell transcriptomics we do not usually have access to direct measurements of protein levels and other elements that take part in the regulatory process. Therefore the complexity of that process is not entirely captured by the *cellSCM* and is instead modelled in a simplified way by a function mapping the expression of the TF-coding gene, and some noise, to the expression of the target gene. In addition, while the true process involves a sequence of events happening one after the other, the *default cellSCM* abstracts time away.

There are different types of biological processes through which genes may impact other genes' expressions, such as transcription regulation depicted in Figure 1.B, but also histone modifications – which control DNA accessibility –, for instance. This means that different edges in the SCM graph may reflect different biological processes, or a combination of them. In fact the SCM graph should not be confused with a Gene Regulatory Network (GRN), even though the two are closely related. The cellSCM is a generative model of expression data, whose underlying graph should capture any causal dependency between observed expression levels. A GRN is a knowledge base used by biologists to organise and summarise known transcription regulation interactions. An overview of these differences is provided in Table 1. Nodes and edges represent different things in each case, and as such, the two graphs are not expected to have the exact same structure. Overall, we expect the edges of the GRN to be a subset of the edges of the cellSCM graph. Table 1 also highlights other differences related to cycles and time modelling which appear as limitations of the *default cellSCM* presented above. Several strategies exist to tackle these limitations, later discussed in sections *Latent causal variables* and *Learning causal kinetic models*.

**Biological perturbations as causal interventions**

A biological perturbation refers to any external action that induces a change in the transcriptomic profile of a cell. Examples of common perturbations include environmental perturbations (pH, temperature changes, signals coming from other cells), but the most common ones are drugs and gene knockouts[40]. Gene knockouts are based on the enzyme CRISPR-Cas9[41]. Guide RNAs (gRNA) are introduced into a pool of cells to target specifically one regulatory region of a specific gene, therefore aiming at modifying a single regulatory mechanism. In the case of a chemical perturbation like a drug, the physical perturbational process depends on the mechanism of action (MOA) of the drug, which is usually summarised by a list of molecular targets to which the drug binds. In general, chemical perturbations are a richer family of perturbations since they can target any type of mechanism, not just transcription regulation.

|  | Graph of the default cellSCM | GRN |
|---|---|---|
| A node represents: | The amount of RNA of a gene at acquisition time (modelled as a random variable). | The abstract idea of a gene (does not take a numerical value). |
| An edge from A to B represents: | Random variable A has a causal effect on variable B.<br><br>This causal effect is the sum of the effects of all the biological processes from gene A to gene B which are not mediated by another gene. This might include transcriptional regulation, as well as other processes (e.g. genes involved in histone methylation have an effect on chromatin folding and therefore on the accessibility of other genes). | Transcriptional regulation: gene A codes for a transcription factor that directly regulates the transcription level of gene B. |
| Contains cycles: | No. | Yes (even self loops). |
| What about time? | Abstracted away. | The GRN describes biological knowledge which is true at all times ("facts", e.g. gene A downregulates gene B during the differentiation of Progenitor X into cell type Y). |

**Table 1**: Overview of the differences between the graph of the *default cellSCM* and a GRN

By viewing single-cell transcriptomics through the lens of causality, it becomes apparent that biological perturbations can be modelled as causal interventions. The most basic assumption is to consider that biological perturbations are perfect causal interventions, as illustrated in Figure 1C. Those perfect interventions remove the dependency of the intervened causal variables on their causal parents. In the case of CRISPR interventions, the level of expression can simply be set to zero to account for a total loss of function of the targeted transcript, while in the case of drugs, the effect could be predicted from the drug structure or other drug features.

However, while the common assumption is that CRISPR gene knockouts are perfect interventions with known targets, it is proven that they carry off-target effects[42–45] and that the perturbational procedure entails uncertainty regarding the success of the perturbation. Also, the standard procedure is to use more than one gRNA per target gene, which leads to a diversity of perturbational effects over the same gene (from null effect to different strengths of the perturbation, along with different off-target effects as well), since each gRNA binds to a different DNA regulatory sequence. Drug perturbations cannot be mapped either to perfect interventions with known targets, since their MOA is not always known, and in many cases, the drug does not directly affect transcription regulation mechanisms. On top of that, there are no guarantees that, for instance, a CRISPR gene knockout or a chemical perturbation entirely breaks the dependency of a target gene to its transcription factor. Finally, biological perturbations do not directly modify the level of expression of the target gene but its transcriptional rate, which later in time leads to a change in its level of expression.

Figure 1.D presents a more realistic view of biological perturbations through a more complex interventional model. Here, biological perturbations are illustrated as non-perfect interventions with off-target effects on other genes, partial removal of the causal dependencies, and uncertainty. Such interventions do not mutilate the graph anymore by removing edges, but modify the structural assignments (between gene G4 and its causal parents gene G2 and G5) by altering their parameters. Overall, the specific impact and nature of these interventions largely remains unknown, which must be taken into account in the causal modelling. For instance, one may rely on different modelling choices and assume that the out-going edges of intervened nodes are affected, instead of in-going edges. To conclude, a simple view of biological perturbations as perfect interventions might fail since it neglects the complexity of the interventional process.

**Independence of biological factors of variation**

The *default cellSCM* assumes that the data generative process is decomposable into a set of independent mechanisms governing the levels of expression. For instance, in Figure 1.A, the mechanism which assigns the level of expression of G2 from the levels of expression of its parents G1 and G3, is independent of any other mechanism in the SCM, in the sense that it is defined by its own function $f_2$, and that interventions can be applied on it without affecting other mechanisms. This is formalised by the Independent Causal Mechanisms (ICM) Principle[17,46,47], which states that the causal generative process of a system is composed of autonomous mechanisms that do not inform or influence each other. Let us try to investigate, based on insights from cell biology, whether we expect the ICM Principle to hold in the *default cellSCM* presented above, i.e. whether we expect the mechanisms governing the levels of expression to be independent.

Each TF protein binds to some specific regulatory DNA sequence, potentially forming a complex with other TFs regulating the same gene, promoting or blocking the recruitment of RNA polymerase for that gene. The rate at which RNA polymerase is recruited defines the rate of transcription of the gene, i.e. the number of RNA transcripts synthesised per unit of time. This shows that for a TF protein to have an impact on the transcription rate of a gene, it needs to be near the promoter region. Note that we are talking about proximity in physical space, not along the DNA sequence – the sequences of TF-coding genes might actually be very far along the DNA sequence. Therefore the mechanism controlling the transcription rate (or *regulatory* mechanism) of a gene is localised in space in the vicinity of its promoter region, and can be considered, in first approximation, independent of the regulatory mechanisms of all other genes, which are localised somewhere else within the cell – close (in physical space) to the promoter regions of other genes. That is to say, if one alters, by applying a surgical and precise intervention, the function that controls the transcription rate of a specific gene given the TFs (which can in principle be achieved via a CRISPR knockout), the functions that control the transcription rates of all other genes should not change since they are localised somewhere else in space.

| | Explanation | Notation and typical assumption |
|---|---|---|
| Regulators | Set of TF coding genes which regulate $g$ (for a given gene $g$). | $reg(g)$ |
| Transcription rate/production rate | Number of RNA transcripts synthesised per unit of time (for a given gene $g$). | $prod(g, t) := pr_g(\{E(r, t), r \in reg(g)\})$<br><br>Typically depends on the levels of expression of the regulators of $g$ (or the concentrations of the corresponding TF proteins). |
| Degradation rate | Number of RNA transcripts degraded per unit of time (for a given gene $g$) | $deg(g, t) := dg_g(E(g, t))$<br><br>Typically depends on the expression of $g$ itself. |
| Level of expression | Number of RNA transcripts in the cell (for a given gene $g$). | $E(g, t) := expr_g(cellular\ state(\{s < t\}))$<br><br>In general, it depends on the history of the cellular state. Typically assumed to be the transcription rate minus the degradation rate integrated over time:<br><br>$E(g, t) = \int^t prod(g, s) - deg(g, s)\ ds$ |
| Mechanism | Any biological process, usually involving chemical reactions, which governs the value of a quantity of interest. | Examples: $pr_g$, $dg_g$, $expr_g$ |

**Table 2**: Presentation of some quantities of interest in transcriptomics, and typical assumptions made to model them.

This discussion suggests that, at least in first approximation, considering the regulatory mechanisms controlling the transcription rates of the different genes as independent seems to be a reasonable assumption. However, the *default cellSCM* models the levels of expressions, not transcription rates. These two quantities are related yet different. The detailed definition and relationship of level of expression and transcription and degradation rates are described in Tables 2 (see for instance *An introduction to systems biology: design principles of biological circuits*[48] chapter 1.4 for more information on how these quantities are related and modelled in Systems biology). Therefore, should we expect the mechanisms governing the levels of expression, modelled by the *default cellSCM*, to be independent?

It can easily be shown that the ICM principle does not hold in general if one applies the *default cellSCM* presented above to data generated through a temporal process in which production rates can be independently perturbed – as the discussion above suggests is possible. We illustrate this point through a simple experiment, see Figure 2.A. We assume a temporal generative process between three genes $G_1(t) \to G_2(t) \to G_3(t)$, and the expression levels of parent genes influence the transcription rates of their children. We simulate dynamics and expression levels are observed at the time of acquisition. As shown in Figure 2.C, a perfect intervention on gene 1 – modifying its production rate – has a clear impact on the observed distribution of gene 3 given its unique causal parent gene 2.

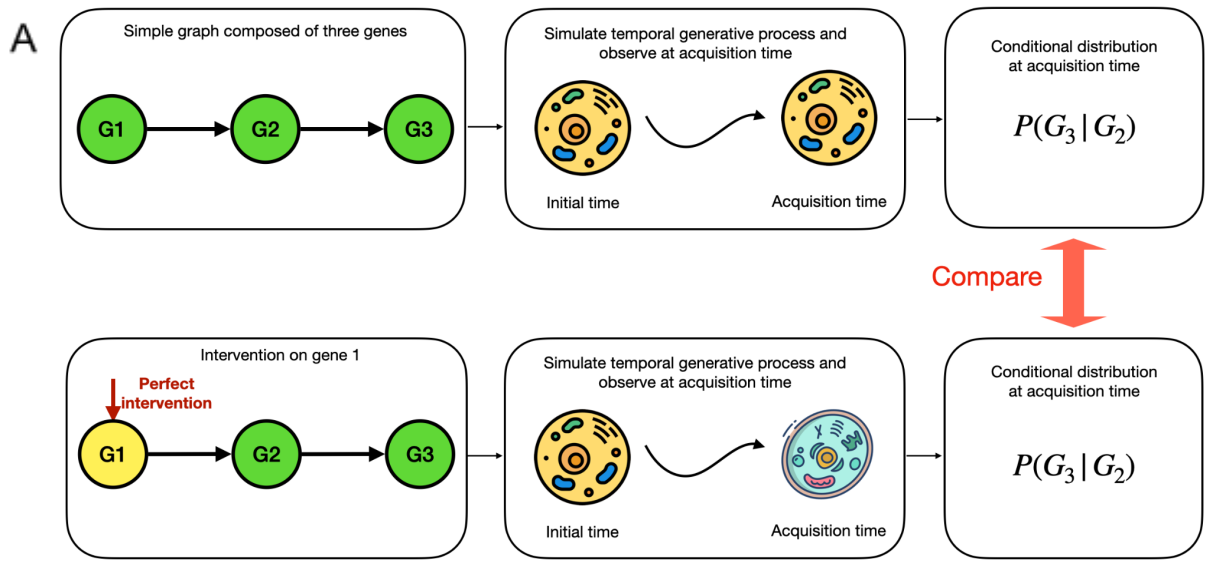

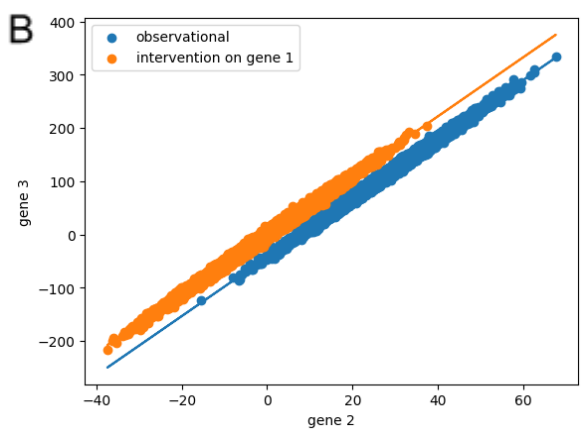
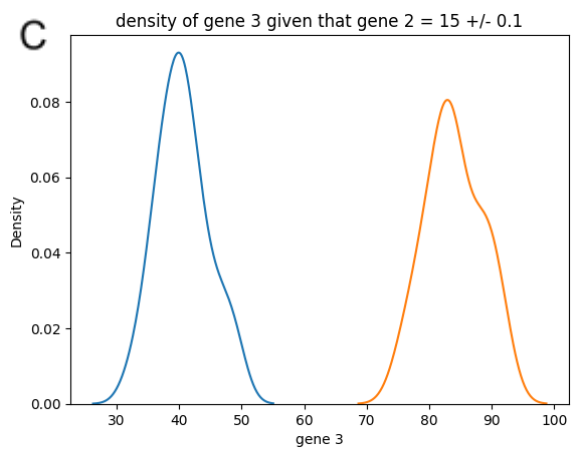

**Figure 2:** The ICM Principle does not hold if one assumes static causal mechanisms while true causal mechanisms happen through time. **(A)** Overview of the synthetic experiment, involving a simple graph $G_1 \rightarrow G_2 \rightarrow G_3$. Dynamics are simulated with and without a perfect intervention on $G_1$, observations are made at acquisition time and resulting conditional probabilities $P(G_3 \mid G_2)$ are compared. **(B)** Observed joint distribution of $G_2$ and $G_3$'s expressions at acquisition time. The function mapping the observed expressions of $G_2$ to $G_3$ has been shifted. **(C)** Observed distribution of $G_3$ given that $G_2$ is equal to 15 (+/- 0.1).

Therefore, in general, the ICM Principle does not hold if one assumes static causal mechanisms and abstracts time away while the true causal mechanisms happen through time. We summarise in Table 3 some cases where we expect mechanisms to be independent or not. In Section *Learning causal kinetic models*, we will see how transcription rates can be directly modelled within the causal model. Finally, the independence between other types of mechanisms, such as the ones governing post-transcriptional regulation or protein degradation could be investigated in a similar manner.

| Mechanisms | Does independence hold? |
| --- | --- |
| Production rate mechanisms (or *regulatory* mechanisms) $pr_g$. | ✅ (in first approximation) |
| Degradation rate mechanisms $dg_g$. | ✅ (in first approximation) |
| Level of expression mechanisms $expr_g$. | ❌ |

**Table 3**: Overview of cases where independence between mechanisms holds or not. Two mechanisms are independent if one can (in principle) modify one without affecting the other.

**Summary**

We have presented a basic approach to causality for single-cell mechanisms. We went over a simple graph-based model consisting of an SCM whose causal variables are the levels of expressions of genes, and a simple interventional model, able to apply so-called perfect interventions. For each of these components, we have tried to explicitly detail what their assumptions entail from the biological point of view, and have challenged these assumptions. We have highlighted some specificities of the *default cellSCM* graph and its differences with GRNs, as well as the complexity of biological perturbations which cannot be understood as perfect interventions. Last but not least, we have shown that one of the main assumptions of the commonly accepted *default cellSCM*, the ICM principle, does not seem to hold in the context of transcriptomics data under this model. Additionally, the default cellSCM has other limitations, mostly related to capturing observed variables only (and disregarding all unobserved ones such as past gene expression levels, protein levels etc), and algorithmic limitations such as its inability to capture cyclic interactions.

Figure 3 illustrates the flaws of the *default cellSCM* when modelling the relationship between a pair of genes G1 and G2. The lower level of expression of G2 at acquisition time $t_{acq}$ can only be explained by the past (and unobserved) activation of G1, and not by the level of expression of G1 at $t_{acq}$.

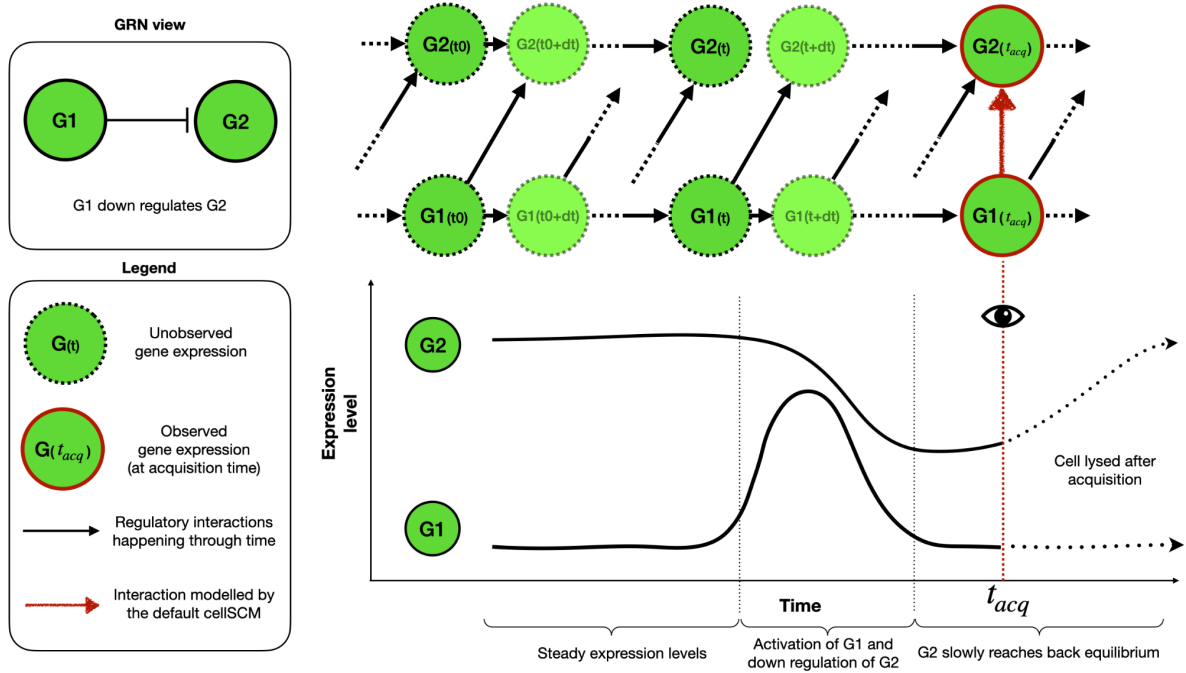

**Figure 3:** Overview of a gene G1 downregulating a gene G2 through time, and how the default cellSCM attempts to model this interaction. Here, G1 gets activated for some external reason before returning to its default expression level. This results in a decrease in G2's expression level, which persists until acquisition time. The observed expression level of G2 cannot fully be explained by the expression level of G1 at acquisition time.

We now introduce three open problems faced by computational methods for single-cell biology and discuss related research directions. We first discuss the challenge of generalisation to unseen environments, and some strategies to build models that are generalisable. We then discuss the interpretability of causal models and its compatibility with the inference of latent variables, as well as the integration of prior knowledge. We finally discuss causal kinetic models, their interpretability and compatibility with cyclic structures in regulatory mechanisms.

# Generalisation to unseen environments

One of the grand challenges of computational biology is to ensure that learnt models generalise to unseen environments, such as unseen cell types or exposure to an unseen drug[49]. All those tasks involve translating knowledge acquired in a certain environment to unseen ones, which is also the goal of out-of-distribution generalisation [50–52] and multi-task learning [53,54], two approaches largely studied in machine learning whose benefits in the context of single-cell genomics has not been clearly demonstrated yet.

We will discuss the importance of the capacity of the interventional model, the difficulty of disentangling between technical noise and biological variation, and the prospective acquisition of informative perturbations.

**Allowed interventions and capacity of the interventional model**

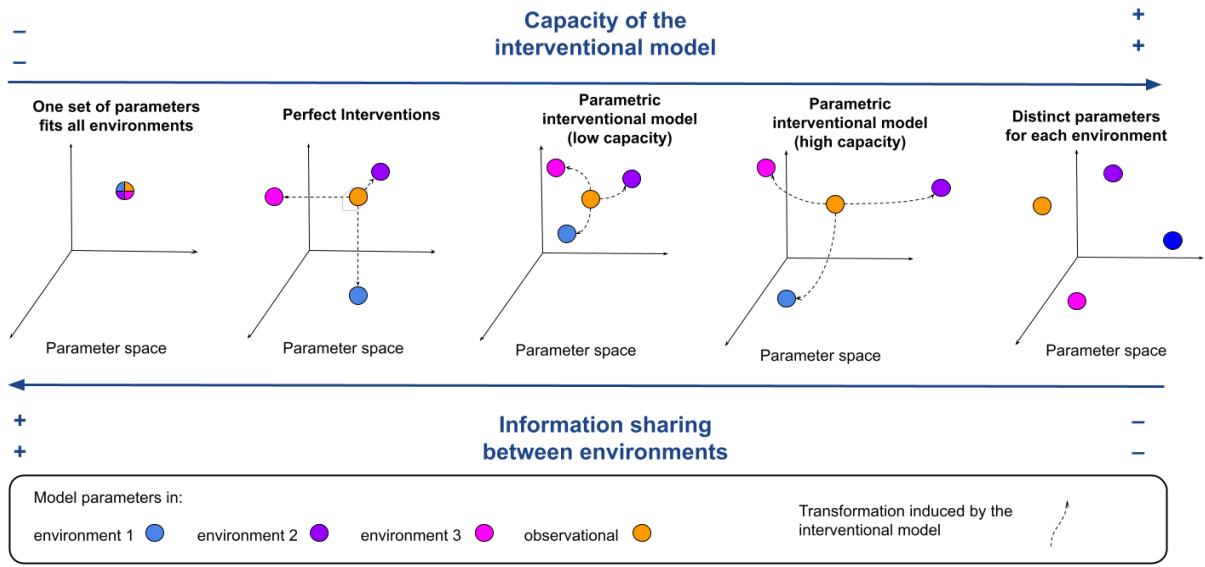

**Figure 4:** Schematic view of several interventional models, ordered by their capacity. Higher capacity leads to a better ability to adapt the graph-based model to each environment, but less information is shared across environments. Perfect interventions are depicted as projections along one of the axis dimensions to illustrate the fact that, in a linear model, perfect interventions correspond to zeroing out one of the rows of the weighted adjacency matrix.

Let us take the example of a causal model that recapitulates the distributions generated by single gene CRISPR knockouts. As discussed previously, modelling these perturbations as perfect single target interventions may be a good approximation if the knockouts have limited off-target effects, but may be an unrealistic way to model perturbations due to the variety of off-target effects that knockouts carry and the uncertainty regarding their success. As a consequence one may decide to consider a broader family of interventional models, such as imperfect single target interventions, or decide to allow for several cellular mechanisms to be modified at once (of the targeted gene, and the off-target effects). At the end of the day, the right choice of interventional model should be based on performance in unseen contexts. These considerations lead us to pay careful attention to the choice of interventional model and its capacity.

There are several ways to formally define the capacity of a model[55]. Loosely speaking, the capacity of a model measures the diversity and complexity of the patterns it can potentially capture. It needs to be adapted to the true complexity of the task at hand in order to avoid the two pitfalls of underfitting – when a simplistic model is unable to recapitulate the complexity of patterns found in the data – and overfitting – when an overcomplicated model captures irrelevant patterns. In causal models, two distinct capacities should be considered: the capacity of the graph-based model $g$, related to the diversity of cells within environments; and the capacity of the interventional model $h$, related to the complexity of changes across environments.

While standard definitions of capacity C may be used for the graph-based model, we propose a specific definition for the interventional model. We call $\theta$ −capacity of $h$ the complexity of the change in parameter space that $h$ can generate given some parameters $\theta$ of the graph-based model $g$. More precisely, let us define the $\theta$ −capacity of $h$, denoted $C(h \mid \theta)$, as the capacity of the following model $\tilde{h}_\theta: I \to \Theta$, where $\tilde{h}_\theta(i) := h(i, \theta)\ \forall i \in I$. Given a prior distribution $p_\theta$ over the parameter space $\Theta$ of $g$, we can define the capacity of the interventional model as $C(h) = E_{\theta \sim p_\theta} C(h \mid \theta)$. A common way to control the capacity of the interventional model is to control the amount of sparsity in the changes it can apply, as in the Sparse Shift Mechanism[56] (SMS) assumption for instance. It states that differences between environments are caused by sparse changes in some of those mechanisms, i.e. only a few parameters differ between environments. The sparser the changes in the parameter space, the more information is shared between environments.

Figure 4 illustrates different choices of interventional models and how they affect the amount of information shared across environments. At one extreme, an interventional model with minimal capacity, which is simply the identity $h(i, \theta) = \theta\ \forall \theta \in \Theta\ \forall i \in I$ results in all environments sharing the exact same parameters. At the other extreme, an interventional model with very high capacity allows one to learn the parameters of the graph-based model independently in each environment.

In summary, causal models feature two capacities that both need to be adjusted: the one of the graph-based model to within-environment variability, and the one of the interventional model to inter-environment variability. However, as we discuss next, data biases may impede generalisation even when interventional model capacity is set appropriately.

**Data biases**

Single-cell data often contains technical noise and measurement errors entangled with the biological signals leading to the so-called batch effects. In addition, biological perturbations can sometimes be unsuccessful. Disentangling these technical variations from biological signals is therefore a key challenge for causal models trained on data coming from multiple batches.

There exists extensive literature on how to account for batch effects and integrate different datasets [13,57–59], however these methods might also remove the biological signals. Regarding the success of the interventions, some methods have been recently proposed to quantify the effect of perturbations[60], identifying perturbed cellular states[61] and discovering cells over which the perturbations have been unsuccessful[62]. The single-cell community effort on data cleaning, integration and improvement of perturbation quality should be leveraged to remove as much technical noise as possible without compromising the biological signal. For example, Aliee et al.[63] incorporate both biological and technical covariates into their generative model and infer conditional prior distributions together with a representation of cells that disentangles biological and technical variations. It is also possible to directly model technical biases into the causal model, for instance as a confounder – a.k.a a variable that influences both parent and target variables. Importantly, the capacity of the confounder – which can be controlled by its dimensionality and the sparsity of its effect – must be adjusted so that it uniquely reflects technical variability and not biological variability.

Experimental replicates – among which all variability is expected to be technical – are a crucial tool to calibrate and validate these approaches. Some types of technical noises could be reduced, or even removed, by standardising experimental protocols between laboratories taking part in the construction of aggregated datasets (or atlases). Other types of noise are harder to tackle, as they are inherent to the context of cell biology, such as uncontrolled genetic mutations, and the stochastic nature of cell differentiation.

**Machine learning-driven experimental design**

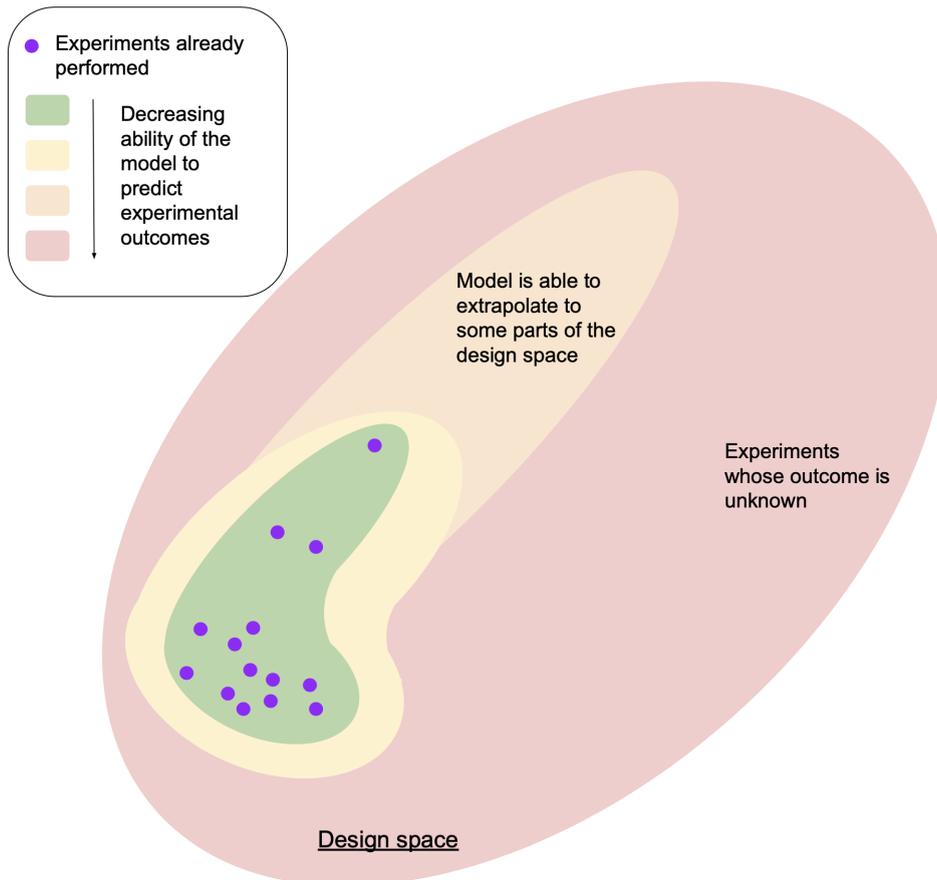

**Figure 5:** Schematic view of model capabilities in different parts of the design space. Given some experiments already performed, a model may be able to extrapolate to some parts of the design space, but might be unable to predict the outcomes of other experiments, if they are very different from the ones already performed.

One of the most promising applications of causal approaches is scientific discovery where experiments are recommended and performed prospectively. However there is usually a large number of degrees of freedom for designing new experiments – together referred to as the design space – and not all possibilities can be tested. Choosing informative experiments requires making some hypotheses about their possible outcomes and estimating the information that they would provide. Instead of relying on human intuition, machine learning models can be used to automate the design of future experiments. In addition, in order to provide maximum power to the pipeline, one is often led to consider large design spaces.

Data corresponding to different environments can be time and cost intensive to acquire. For this reason, it is common to have access to only a limited – and possibly concentrated in a specific region of the design space – set of previously acquired environments. Models trained on such data, whether they are causal or not, might be unable to generalise well to parts of the design space that are too different from the training data, as depicted in Figure 5. Understanding regions of the design space that are well captured by the model (*i.e.* where model predictions can be trusted) is therefore extremely important, especially when predictions are to be used for downstream decision making. To this end, one can estimate epistemic uncertainty, which is related to the lack of knowledge of the model about the world, and which could be reduced if more data were to be available. The traditional approach for estimating epistemic uncertainty consists in estimating the full posterior over model parameters, either explicitly[64] or implicitly[65,66]. Epistemic uncertainty can also be estimated directly via an auxiliary model[67]. Uncertainty estimation has been adapted to the context of causal models in order to obtain a posterior over graph structures [33,65,99].

The next step is to leverage model predictions and epistemic uncertainty within a strategy that can guide the design of future experiments. The challenge is to come up with strategies that are as sample efficient as possible, that is, which can reduce epistemic uncertainty (a case known as Active Learning), or maximise some property (a case known as Sequential Model Optimization or Bayesian Optimization) as fast as possible. In order to recommend future experiments, one has to select candidate experiments, evaluate their predicted outcome, estimate how informative these experiments would be, and select the most promising candidates that will actually be recommended. In the case of extremely large design spaces, sampling good candidates can become difficult, and one has to rely on efficient sampling strategies[69]. These sequential approaches can be adapted to the context of causal models[70,71] in order to recommend informative interventions.

Active Learning and Sequential Model Optimization have shown great promise in various areas of science, including molecular property prediction[72,73] and material design[74]. However, their application to cell-based assays, where batch effects are huge[59], is still very challenging. Some methods have bypassed the batch effect issue by relying on cell-free systems[75], or have focused on yeast and bacteria[76–78] which are relatively stable. In the case of human cells, careful understanding of the biases at hand (including batch-to-batch variability) is needed in order to query informative perturbations. To the best of our knowledge, drug combinations is the only context in which Sequential Model Optimization has been applied prospectively, and quantitatively validated, to human cells[98]. Adapting and scaling up these methods to, say, CRISPR knockout recommendation for GRN inference, is an open challenge.

Machine learning-driven experimental design for cell-based assays holds great promise for efficiently guiding experimentation and accelerating the discovery of cell mechanisms and drugs' mechanisms of action. In many cases however, the lack of interpretability of the models may hinder the full integration of these approaches into the workflow of biologists.

# Learning interpretable models

Ideally, a causal model of single-cell transcriptomics would not only generalise to unseen scenarios by modelling different distributions but also offer an interpretable perspective on the cellular mechanisms. We consider a model – or part of a model – interpretable when the operations it performs can be associated with known processes (e.g. transcriptomic regulation, binding etc) and the values it stores can be associated with known and measurable biological quantities (concentration of a specific molecule, spatial conformation/accessibility of a molecule, etc.). Such an interpretable view would be useful for expanding scientific knowledge and solving problems. For instance, an interpretable causal model that encodes relationships between known biological entities would allow biologists to distil knowledge from the model and incorporate it into their analysis and understanding of cellular systems.

After describing the inference of latent causal variables, which is a common strategy used to overcome partial observability and computational issues, we analyse the challenges it poses to the interpretability of models, and finally describe how prior knowledge can be integrated into interpretable models.

**Latent causal variables and their challenges for interpretability**

In theory, we know with great precision the complete set of biomolecules (all small molecules, all proteins, all RNAs and DNA) found inside cells. However, for the vast majority of biological functions, only a tiny fraction of all biomolecules is involved. Given that working with the complete set of human genes (more than 20k) imposes algorithmic and computational challenges, a useful simplification is to carefully select a small set of relevant causal variables, which depends on the problem at hand. On the other hand, not all variables of interest are observed, due to the partial observability of cells (i.e., our measurements do not recapitulate all the relevant aspects of the cell state), and therefore it is common to allow for non-observed – or latent – variables within the causal model. For instance, chemical perturbations such as drugs might target proteins – or protein complexes – whose concentration is not typically measured in single-cell data. In that case one may choose to model the unobserved concentrations of targeted proteins. Modelling chromatin accessibility with latent variables from expression data would be another example.

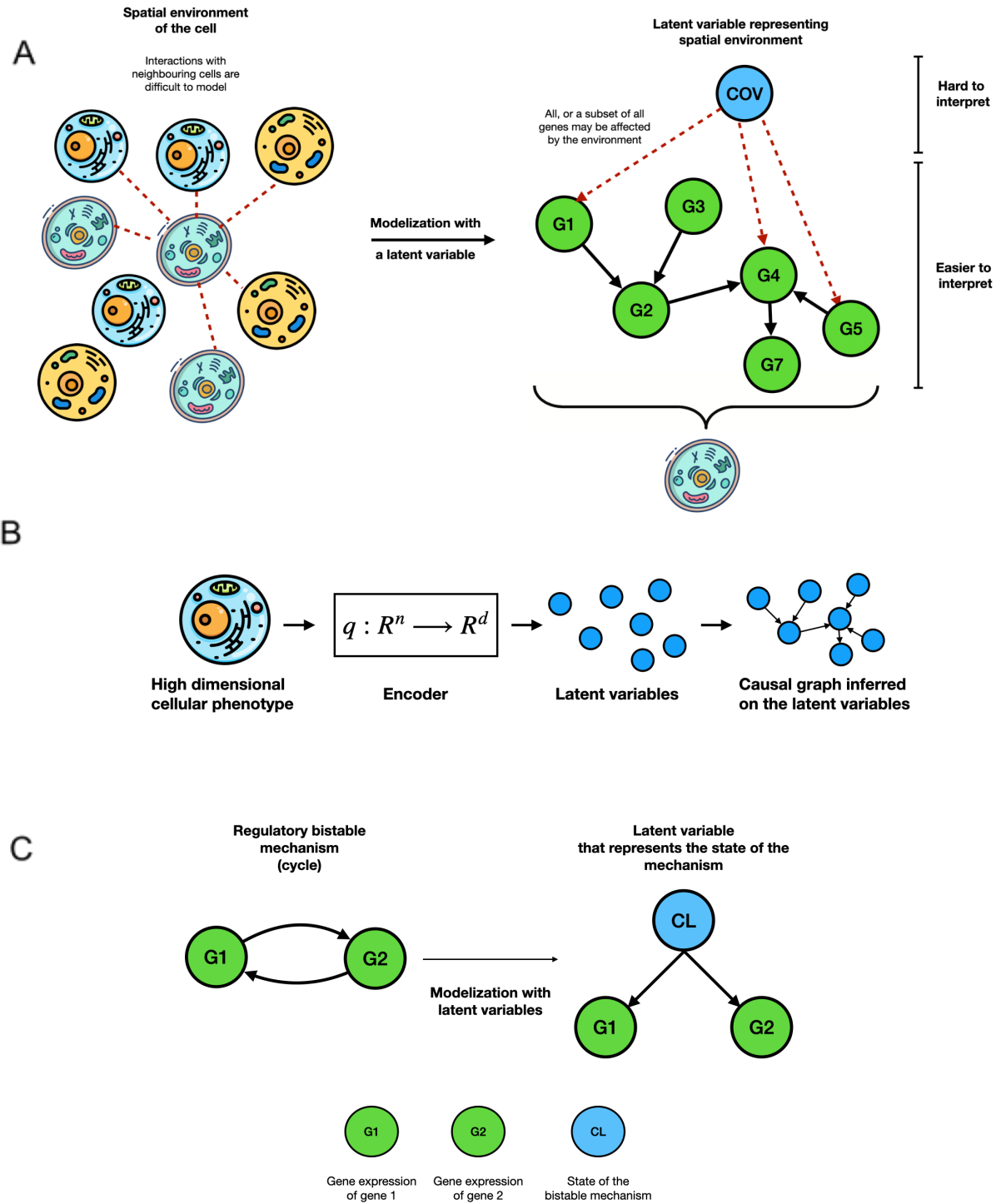

**Figure.6:** Latent causal variables may impede interpretability but may also solve algorithmic challenges **(A)** Complex processes that are difficult to model like the effect of the spatial environment can be captured by latent variables, and added as additional causal variables in the causal graph. Learned latent variables might be hard to interpret. **(B)** Causal representation learning aims at discovering latent causal variables and learning interactions among them. These two tasks can either be tackled sequentially or jointly. The encoder $q$ can be any type of machine learning architecture. **(C)** Operating with latent variables may resolve algorithmic problems like the existence of cycles in the causal graph.

Information about the state of high-level biological structures, conformed by groups of genes or other biomolecules (such as protein complexes) may also be captured through latent variables. In this case one may refer to latent variables as high-level variables. The value of these high-level causal variables would typically be inferred from a collection of low-level observations (genes, etc.). They could in principle correspond to some attributes of known mechanisms such as the activation of gene programs, pathways, as well as messages in cell-cell communication, aspects of the spatial environment, or any other variable that directly or indirectly plays a role in the regulation of expression levels. Figure 6.A illustrates the specific case of incorporating spatial information into a causal model. Modelling all the different interactions of the cell with its spatial environment and the associated processes may be challenging; instead, encapsulating the spatial effects into a covariate variable may ease the problem, at the expense of interpretability.

The identification of low-dimensional single-cell data representations has been widely explored mainly through disentanglement techniques or matrix factorization[79]. The newborn field of Causal Representation Learning[56] (CRL) precisely aims at discovering high-level causal variables from low-level observations. Some works rely on CRL to discover underlying regulatory elements that control cell behaviour[63,80,81]. Much of the existing work on CRL is still based on disentanglement[82–84], which simplifies the problem of recovering meaningful representations with causal knowledge assuming a trivial causal graph among latent variables, i.e., an empty graph with no interactions. In single-cell transcriptomics, that setting seems overly simple. Ideally, a change in one high-level variable should trigger downstream changes in other inferred causal variables. In Figure 6.B, we provide an overview of the CRL strategy, which consists in learning both latent variables and causal dependencies between them.

A notable benefit of latent variable inference is that the value of a latent variable may be less subject to technical noise since it is not a single experimental measurement but is usually derived from several measurements. Furthermore, working with latent variables can solve algorithmic limitations such as the existence of cycles in the causal graph. For instance, Figure 6.C illustrates a potential use case of latent causal variables to solve the existence of cycles in the GRN, a case not contemplated by the *default cellSCM*. Gene 1 and 2 take part in a bistable mechanism and down-regulate each other. This mechanism allows for two stable states where one of the genes is expressed but not the other. Instead of modelling the cycle structure, one can learn a latent variable that accounts for the current state of the bistable mechanism and that has a causal effect on the expressions of the two genes.

Last but not least, learning latent variables significantly impacts the interpretability of the model. Indeed, it is most of the time unclear which biological entities or quantities the latent variables correspond to. Even though the idea of directly mapping gene programs or other known biological structures to inferred causal latent variables is appealing and could enhance interpretability, it is challenging and until now no significant results have been achieved. It is important to note that a causal model may consider both fine-grained and coarse-grained variables, i.e., one may operate at the gene level but also model the spatial environment as a single coarse-grained variable instead of individually modelling all elements of the spatial environment.

As a summary, learning latent variables offers many advantages to deal with the partial observability of cells, as well as computational limitations of models which rely on observed variables only. However, this strategy significantly hinders interpretability, and reconciling latent variable inference with interpretability is a challenge.

**Prior knowledge addition**

Years of research in cell biology and systems biology have distilled a vast amount of prior knowledge that can be incorporated into models to complement the information contained in the data. Modelling cellular processes through a graph-based model enables a straightforward path to prior knowledge addition into causal models (when the prior knowledge comes in the form of a graph or multi-graph). However the integration of such knowledge is only possible in an interpretable model, where causal variables can be associated with the known biological entities found in the prior-knowledge database.

Prior knowledge integration may offer many benefits. For instance, it may ease the task of complete graph discovery by providing a good prior or initialisation to the causal discovery algorithm, or turning the task into a partial graph discovery[33]. For example, there exists prior knowledge regarding known connections between transcription factors and target genes. Prior knowledge can also provide information about the biological functions a gene is involved in (e.g. cell cycle). Groups of genes known to function together, such as gene programs or biological pathways, are referenced in databases[85]. Assessing the quality of available prior knowledge, and figuring out the extent to which they can improve model performance (in which context, and under which model assumptions) is an ongoing challenge[86].

# Learning causal kinetic models

In models discussed so far, time has not been explicitly considered. Yet, biological mechanisms happen through time, and in many cases, such as in cell differentiation, the temporal aspect cannot be ignored. In addition, models that abstract time away may rely on hypotheses that do not hold in practice due to the dynamical nature of biological processes, such as the *default cellSCM,* as we saw in Section *Independence of biological factors of variation*. Taking time into account in the causal model might allow us to rely on other hypotheses, more in line with the real nature of biological processes.

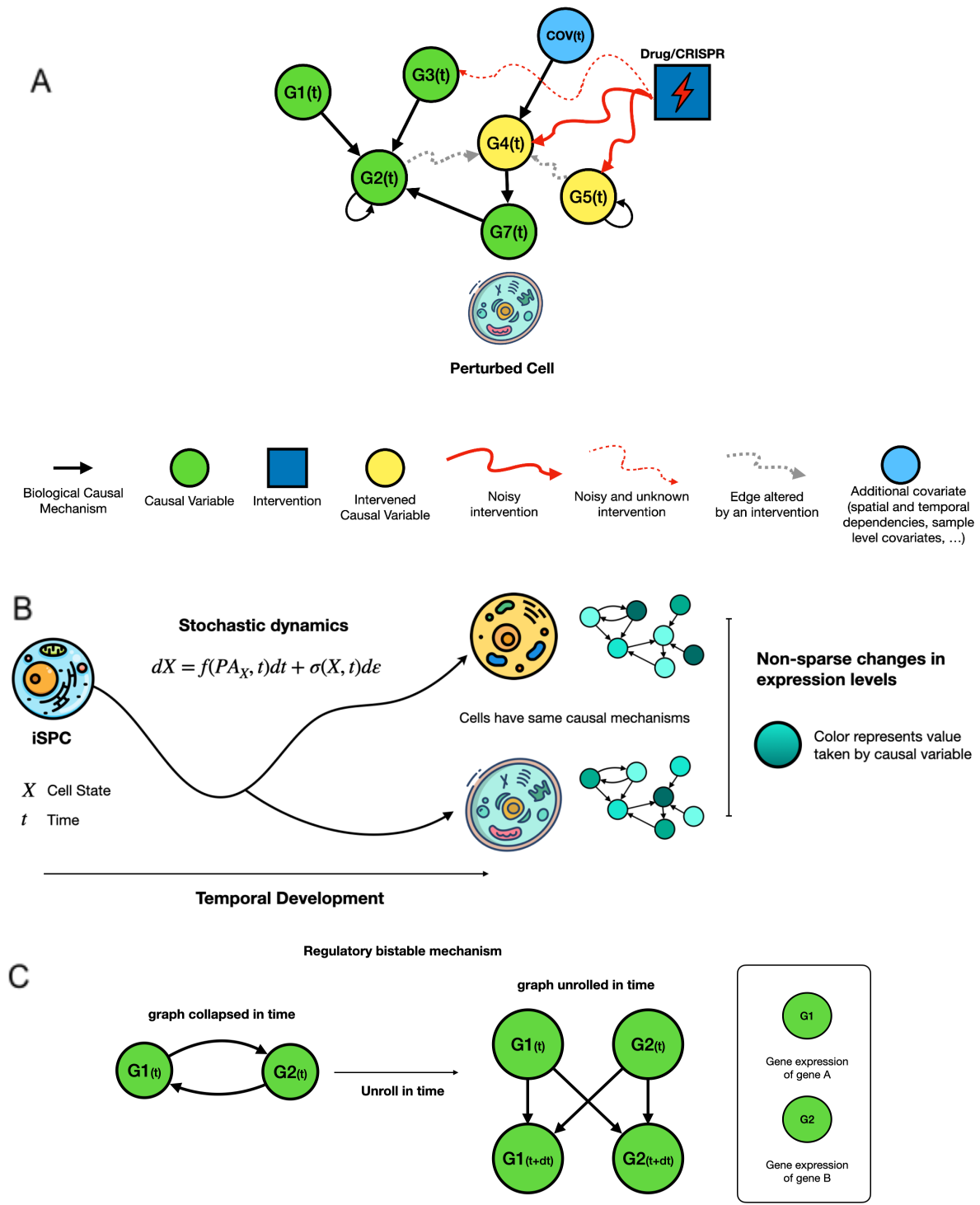

**Figure.7:** Cells can be modelled through causal kinetic models **(A)** Overview of a causal kinetic model for cells. Variables represent expression levels and now explicitly depend on time. Their evolution is a function of their causal parents. Perturbations may be applied in the same way as before, and latent variables may be learnt. The graph may now contain cyclic regulatory motifs and autoregulation. **(B)** Cellular development is a dynamical process which can be modelled by differential equations, either ordinary or stochastic. Due to stochasticity, branching may happen and cells driven by similar mechanisms may end up with different expression profiles. **(C)** The graph collapsed in time may present cycles, which are incompatible with static SCMs. In causal kinetic models, cycles no longer pose algorithmic difficulties, as each variable is a function of its parents and itself, evaluated in the past, as shown in the graph unrolled in time.

In SCMs relationships are drawn between the static values of the causal variables and structural assignments directly govern the values of causal variables. In contrast, causal kinetic models natively incorporate temporal information and account for the dynamical properties of the system under study. In causal kinetic models introduced by Peters et al.[36], structural assignments govern the rate of change of causal variables by relying on ordinary differential equations (ODE) and stochastic differential equations (SDE), two of the most popular mathematical tools for modelling temporal evolution of systems. More precisely, the rate of change of each variable only depends on the value of its parents, and itself, potentially. This means that the evolution function is assumed to have sparse dependencies that correspond to the edges of the causal graph. Interventions may be applied in a similar way as before by replacing one (or several) of the structural assignments, as shown in Figure 7.A.

Relying on differential equations could result in a sparser causal graph compared to modelling causal dependencies across long periods of time. Otherwise, indirect effects (mediated through multiple time steps, each of which involves only a sparse directed graph) would show up as direct causal links in the causal graph. Mathematically, this can be seen as follows: multiplying two sparse matrices yields a less sparse matrix. Hence, the repeated application of a per-time-step causal dependency structure corresponds to a resulting causal dependency graph whose adjacency matrix is associated with the per-time-step adjacency matrix taken to some power (the number of time steps). The resulting graph may therefore be highly connected (non-sparse) when considering causal dependencies across long periods of time. Another advantage, specific to the stochastic version of causal kinetic models based on SDEs, is their ability to account for branching phenomena observed in practice in cell differentiation trajectories, see Figure 7.B.

There exist two views of the causal graph in the dynamic setting: the graph collapsed in time and the graph unrolled in time, as shown in Figure 7.C. The graph collapsed in time is a graphical representation of the dependencies between variables and their time derivatives, as defined by the evolution function. The graph unrolled in time is a discretization of the continual evolution processes modelled by the dynamical system, i.e. a node now corresponds to the expression of a gene evaluated at a given time t. While in the graph unrolled in time there exists no cycles since instantaneous changes are not allowed, cycles can exist in the graph collapsed in time. For instance, a causal variable can be a causal parent of itself, a motif known as autoregulation, creating a loop (in the graph collapsed in time). Along the same line, there can exist other cycles of higher order (involving two or more variables) in the graph collapsed in time. These kinds of loop motifs are common in the regulatory connections between genes, and often serve as feedback loops which keep the system at equilibrium. In conclusion, cyclic structures cannot be captured in a static setting, but do not pose any difficulty when cells are seens as a dynamical system.

The main difficulty in the application of these methods to single-cell data resides in the snapshot nature of the data: single-cell sequencing is a destructive procedure wherein cells are destroyed when measured. This means that these models attempt to model dynamics which are only observed at one time point – the time of acquisition. One strategy is to rely on pseudotime inference methods[87–89], which associate each single cell with a different pseudotime recapitulating the extent to which the cell is differentiated, i.e., its location along its differentiation branch. Dynamical models have been applied to single-cell data by relying on pseudotime[90,91]. Alternatively, several experiments can be performed with different incubation times. This allows for the analysis of trends at the population level, but there exists no explicit matching between cells at different time points. In order to tackle this problem, there exists an effort in the single-cell community for inferring the trajectory of individual cells through Optimal Transport[92–95]. Interestingly, recently developed techniques can keep cells alive after transcriptome profiling, but these experiments are not scalable enough yet[96].

Causal kinetic models offer a rich description of biological processes and have several advantages compared to their static counterparts. For instance, causal kinetic models can be used to generate trajectories under different interventions, opening the door to investigations of the different developmental trajectories that cells follow from the primitive states (iSPC) to fully developed functional cells. While an intervention in a causal kinetic model may generate subtle effects in the short term, it can lead to major ones in the long term due to the progressive propagation of causal effects to all downstream variables, making it challenging to understand and identify the mechanisms at play. While the problem of identifiability – that is to say whether we are able to discover the true parameters and mechanisms of the system given enough observations across enough environments – has been studied in the context of Granger causality[97], it has barely been studied in the context of causal kinetic models. To make progress in this direction, we believe in the crucial importance of having access to interventional temporal data. Most of the current perturbational assays simply measure control and perturbed conditions after a given incubation time. In that two-time-points setting, learning causal dynamics can be challenging. More perturbational assays with multiple temporal measurements of control and perturbed cells populations are needed.

In conclusion, causal kinetic models are a valuable tool for causal approaches in single cell transcriptomics. They are consistent with the temporal nature of biological processes, and offer a simple way to overcome some of the difficulties encountered by static methods, such as the existence of cycles. They might also lead to additional interpretability as they allow for the prediction of temporal dynamics. However, they also face great challenges, mostly related to the scarcity of time points at which observations are made. Note that the previous concerns regarding interpretability and generalising to unseen environments still hold. Careful attention to the capacity of the interventional model should still be paid. Likewise, one may combine high-level and low-level causal variables and incorporate prior knowledge or multi-omics layers.

# Conclusions

In this perspective we have investigated how the framework of causal machine learning applies to single-cell genomics. We have introduced the classical causal view of a cell and discussed its assumptions and limitations from the biological point of view – taking into account the specificities of single-cell data and its regulatory mechanisms. Finally, we have identified and discussed three open problems: generalising to unseen environments, preserving the interpretability of models, and learning causal kinetic models.

Figure 7.A illustrates our final view of a causal model for single-cell biology. It is not acyclic, since it describes a dynamical system that evolves through time with cycles and autoregulation. The interventions are not perfect anymore, and they entail off-target effects and technical noise. It contains low-level variables and latent variables. Strategies to learn such a model include experimental design techniques that efficiently explore huge design spaces. Ideally, careful attention should be paid to the way high-level variables are inferred, in order to preserve the interpretability and allow distilling the learnt model into scientific knowledge. Nevertheless, the causal perspective provided here still faces unsolved challenges.

First, without access to ground truth data to evaluate results, and without identifiability guarantees, derived from the use of noisy real-world data, causal approaches need to be tested in some downstream task and experimentally validated to prove their veracity. In general, if a causal approach does not provide good performance in unseen environments, we do not know whether the assumptions imposed on the model are faulty, or whether the assumptions are accurate but the inference procedure is faulty (i.e. it does not infer the right graph structure – or posterior over graph structures – and parameters). Distinguishing between these two cases would require evaluating causal approaches on data where model assumptions are known to hold, which could be generated through simulations. Second, the described model is still susceptible to the effect of unknown confounders due to the partial observability of the system. Constant improvements in single-cell sequencing technologies enable ever richer observations of the state of single-cells, potentially alleviating the confounder problem.

Causal approaches offer the hope to model the mechanisms that rule the behaviours and dynamics of complex systems such as cells. However, this is the very early age of causal machine learning applied to single-cell biology, and while many methods have been presented in this perspective, a large effort remains to be made to experimentally validate them, which has been challenging so far due to the relatively lack of perturbational data, a key pivot for causal discovery. Fortunately, continuous advances in experimental technologies for single-cell sequencing should keep increasing the availability and quality of data.

**Code availability**

Our code is available online at
https://colab.research.google.com/drive/1JiGwxqPrIT7yiAm0iHToub5VoVvgIY3h?usp=sharing .